\documentclass[letterpaper, 10 pt, conference]{ieeeconf}
\IEEEoverridecommandlockouts
\usepackage{cite}
\usepackage{amsmath,amssymb,amsfonts}
\usepackage{algorithmic}
\usepackage{graphicx}
\usepackage{textcomp}
\usepackage{xcolor}
\usepackage{soul}
\usepackage{subcaption}
\usepackage{tabularx}
\usepackage{booktabs}
\usepackage{makecell}
\usepackage{siunitx}
\usepackage{dblfloatfix}    
\usepackage{hyperref}
\usepackage{url}

\def\BibTeX{{\rm B\kern-.05em{\sc i\kern-.025em b}\kern-.08em
    T\kern-.1667em\lower.7ex\hbox{E}\kern-.125emX}}
\begin{document}

\title{\LARGE \bf
Human-Guided Co-Manipulation of Carbon Fiber Plies
}

\author{Rami Ojanen$^{1}$, James Fant-Male$^{1}$ and Roel Pieters$^{1}$
\thanks{$^{1}$Automation Technology and Mechanical Engineering, Tampere University, 33720, Tampere, Finland, {\tt\small firstname.surname@tuni.fi}}%
}

\maketitle
\thispagestyle{empty}
\pagestyle{empty}

\begin{abstract}

The handling of flexible materials is a difficult task to fully automate due to the challenges caused by the deformability of these types of objects. Meanwhile, a fully manual process can be ergonomically challenging, tedious and inefficient. Thus, human-robot collaboration (HRC) and cooperative manipulation (co-manipulation) have received increasing interest in this field as they enable human involvement when needed while also improving productivity. To enable efficient co-manipulation and interaction between the human operator and the robot, different modalities and control methods are required. In this paper, we present and examine different control methods for co-manipulation of carbon fiber plies, evaluating the pros and cons of each method in a controlled setting. We propose that a multimodal combination of
speech commands, wrist-tracking through vision, and force with compliant control would provide the best solution for complete and intuitive control of the task.

\end{abstract}


\section{Introduction}

The manipulation of composite materials, such as carbon fiber sheets, and other flexible materials, is an important task required in various manufacturing settings, for example, in the aviation and automotive industries~\cite{bjornsson_automated_2018}. Due to certain challenges in robot cognition and manipulation, these tasks still lack fully automated solutions and are mainly performed manually~\cite{makris_deformable_2022}. A completely manual process can be ergonomically challenging, tedious and inefficient~\cite{malhan_automated_2021}. Human-Robot Collaboration (HRC) and human-robot cooperative manipulation (co-manipulation), which have gained increasing interest in recent years, present a promising solution. They allow human involvement in the process for parts that require human participation, such as decision-making, while also enhancing efficiency and productivity~\cite{bonci_human-robot_2024}.

To enable effective and safe co-manipulation of flexible materials and close collaboration between human and robot, interaction and communication between them are essential. As the operator needs both hands during the manipulation process, using physical interfaces such as a teach pendant to control the robot is usually not feasible. Additionally, due to a dynamic manufacturing environment and varying process requirements, a robot that simply follows a predefined trajectory is rarely effective. Thus, utilizing different control methods and modalities, such as speech, vision and force, is an attractive option. Voice commands can be an intuitive way to control the robot, while force and compliance allow the human operator to easily, safely and fully control the manipulation process by pulling or moving the robot or material being manipulated. Meanwhile, vision allows the robot to follow the user without directly applying force, which can be advantageous for both the robot, human and the material being manipulated.

\begin{figure}[t]
  \centering
  \includegraphics[width=\linewidth]{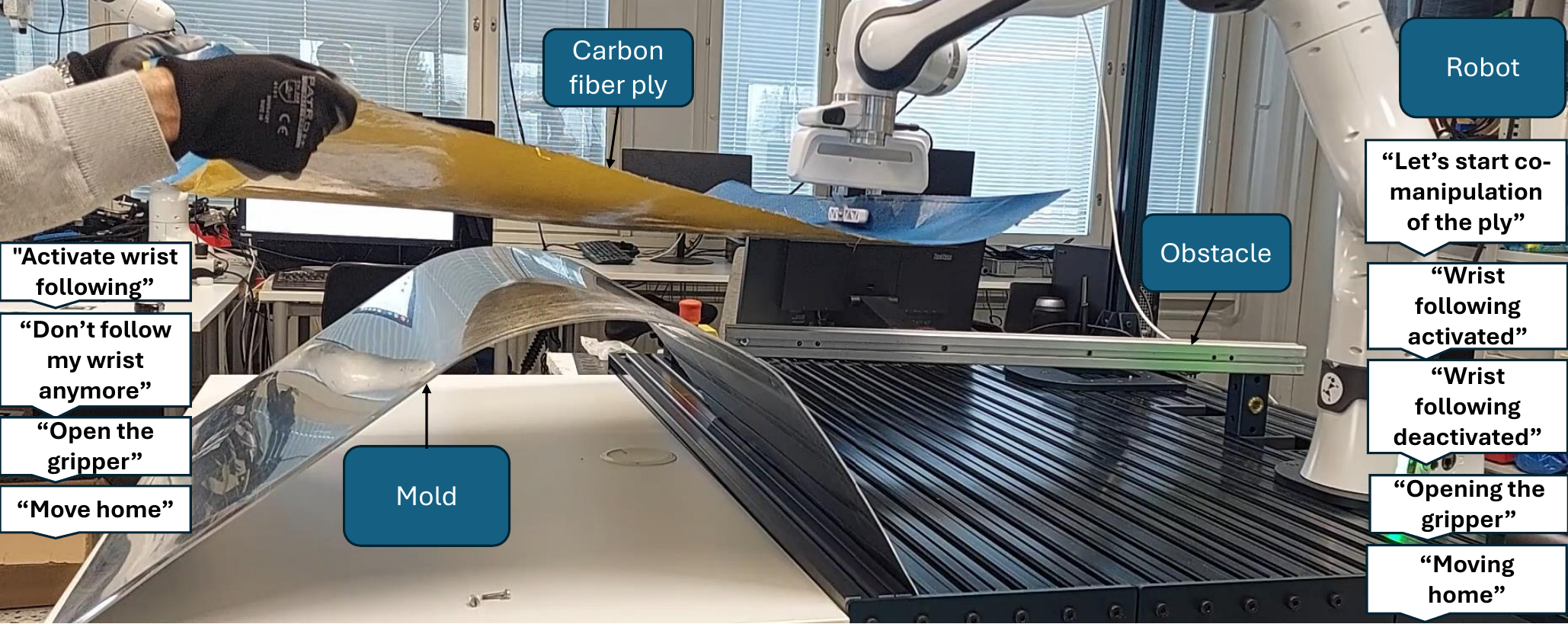}
  \caption{User co-manipulates carbon fiber ply with robot. Control options include voice commands, compliance, wrist tracking and a predefined trajectory.}
  \label{fig:overall}
\end{figure}

Although various strategies have been explored for the co-manipulation of flexible materials in different settings~\cite{kruse_collaborative_2015,de_schepper_towards_2021,bonci_human-robot_2024}, there is no clear comparison between approaches. In this paper, we present and analyze different control methods for co-manipulation of flexible carbon fiber plies, including compliance, a predefined trajectory, step-by-step voice commands and wrist tracking. Additionally, this work proposes how a multimodal combination of speech, vision and force could be used in this task.

The rest of the paper is organized as follows: Section~\ref{sec:related} contains the related work, Section~\ref{sec:use_case_description} describes the use case and our setup, Section~\ref{sec:methodology} presents the methodology with the different modalities, how they were implemented and utilized. Section~\ref{sec:experiments} describes the experiments performed and results, which are then discussed in Section~\ref{sec:discussion}. Finally, Section~\ref{sec:conclusion} concludes the paper.

\section{Related work}
\label{sec:related}

\subsection{Human-Robot Collaboration and Interaction}

Developing human-centric manufacturing systems is a key part of Industry 5.0, prioritizing humans at the center of production systems and aiming to increase synergy between humans and machines~\cite{wang_toward_2022}. This has led to increasing interest towards HRC and Human-Robot Interaction (HRI). The aim is to improve the productivity, flexibility and efficiency of factories as well as reduce the load on human operators by combining the intelligence and flexibility of humans with the accuracy and consistency of robots~\cite{inkulu_challenges_2022}. HRC and HRI have applications especially in manufacturing~\cite{wang_multimodal_2024}, for example in assembly and inspection tasks, while in healthcare, social robots can help patients and ease the workload of hospital workers~\cite{ragno_application_2023}.

Multimodal communication and control strategies are essential to ensure effective collaboration and interaction. Typical modalities of HRI include vision, auditory and language, physiological sensing and haptics. Common examples of vision include object detection, human body recognition and intention prediction, while automatic speech recognition (ASR) and text-to-speech (TTS) systems allow verbal communication between humans and robots. Meanwhile, devices like EEGs (Electroencephalography) and IMUs (inertial measurement unit) allow physiological sensing, and force control and tactile sensing are common examples of haptics. \cite{wang_multimodal_2024}

\subsection{Co-Manipulation of Flexible Materials}

Manipulation of deformable and flexible materials is a very challenging task for a robot to perform alone. The dynamic distortion and changing state of this type of material impose several challenges in cognition and manipulation for robots~\cite{makris_deformable_2022}. Thus, human-robot co-manipulation, an important area within HRC, has also received interest in this application area. Human involvement in decision-making and manipulation, especially when handling bigger sheets of material, can be advantageous.

Different approaches have been proposed to enable effective co-manipulation of deformable objects and flexible materials. Common approaches include force- and vision-based methods, and different hybrid combinations of them. Other options include model- and learning-based methods~\cite{makris_deformable_2022, 8769898}, which can be computationally heavy and sensitive to changes in the environment. In~\cite{liao_trust-based_2024}, an object variable impedance control with a variable stiffness law based on trust and force-based human intention estimation is presented. With camera-based methods, usually either the state of the material~\cite{10309337} or the location of the human operator~\cite{bonci_human-robot_2024}, through e.g. skeleton tracking, are monitored and this information is utilized as part of the control strategy.

With hybrid force-vision controllers, the aim is to combine the advantages of both methods. In~\cite{de_schepper_towards_2021}, a controller using sensor-fused force-torque and skeleton tracking data for mobile co-manipulation is presented, while in~\cite{kruse_collaborative_2015} vision is used to detect wrinkles in the ply and generate a corrective vector that is combined with the force controller. Meanwhile, in~\cite{chen_human-guided_2024}, real-time body tracking and impedance control are combined to create a control paradigm for changing between different interaction modes in a co-manipulation task.

The methods in the literature present important developments in different control strategies for handling co-manipulation tasks. However, there lacks a clear comparison of methods and how they differ in a co-manipulation task. This motivates this work, where we focus on comparing key approaches to co-manipulation: voice commands, vision based wrist-tracking, compliant control and a hybrid combination of those, with comparison against a predefined trajectory method. We focus on methods that do not require modeling of the material itself, given the added computational complexity and reduced transferability when deploying a solution to new scenarios.

\section{Use case description}
\label{sec:use_case_description}

Our developments target an aircraft seat manufacturing use case, though our methods and findings are applicable to similar tasks where large composite sheets must be placed on a mold as part of a layup process. 
Currently, the process is labor-intensive with all tasks performed manually. In the case of handling large plies, two workers are required to place the material on the mold effectively. By integrating collaborative robots and other technologies into the task, the objectives are to reduce operator effort and exposure to unhealthy materials, and to improve the quality and efficiency of the process.

In this work, we focus on the transportation and manipulation of the ply through human-robot co-manipulation. Different control methods and modalities for this task are examined and compared in two different cases: with and without an obstacle. Our setup consists of a simplified HRC workcell where the pros and cons of different control strategies can be analyzed in a controlled setting, see Fig.~\ref{fig:overall}.
We use prepreg carbon fiber plies without removing the backing film, and a simplified aluminium mold to place the ply over during the co-manipulation process. An aluminium profile bar on top of small blocks is used as an obstacle. The obstacle adds complexity and variety to the experimental setup, forcing more complex robot trajectories already while approaching the mold. Having obstacles in the scene is also common in related work~\cite{liao_trust-based_2024,villagrossi2025efficient}.

\section{Methodology}
\label{sec:methodology}

\subsection{System Architecture Overview}

\begin{figure*}[t]
  \centering
  \includegraphics[width=1.0\textwidth]{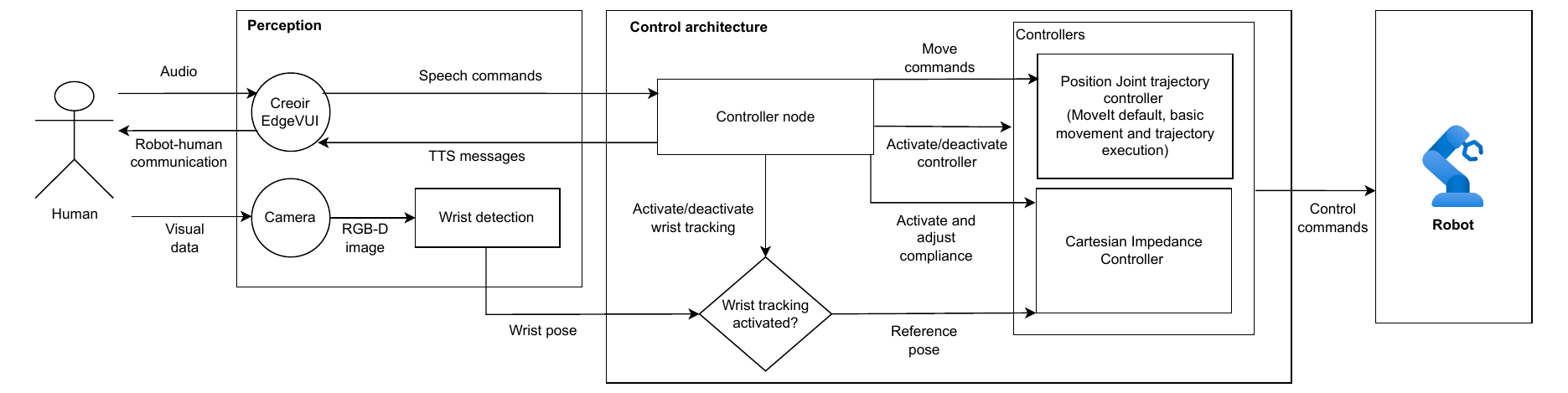}
  \caption{Overview of the system architecture illustrating data flow between modules.}
    \label{fig:architecture}
\end{figure*}

An overview of the system architecture is presented in Fig.~\ref{fig:architecture}. It consists of three main blocks: perception, control architecture and the robot. The perception block receives audio and visual data from the environment, processes it and transmits the information to the control block as speech commands and wrist poses. It is also responsible for producing robot-human communication through TTS functionality. The implementation and utilization of vision, and both speech recognition and TTS are explained in more detail in the following Subsections C and D, respectively.

The controller node is the main part of the control architecture. It integrates audio communication by connecting the recognized speech commands to corresponding actions and sends messages for the TTS functionality when robot-human communication is needed. It also activates and deactivates the required control methods, for example wrist tracking or compliance, and controllers, such as Cartesian impedance controller, when necessary. It communicates with the gripper, MoveIt~\cite{moveit_case_study_2014} and the controllers, sending commands for movement and grasping or messages for adjusting the compliance, which are then responsible for the lower-level implementation and communication of these actions with the robot. A Franka Research 3 (FR3) collaborative robot was used, and our open-source implementation targeted for ROS2 Jazzy is available at \url{https://github.com/ramblam/Human_Guided_Co_Manipulation}, including videos of the experiments.

\subsection{Impedance Control}

The ability of the robot to interact with its environment in a compliant manner allows the human operator to easily control the manipulation process by pulling or moving the ply while the robot adapts to human guidance. Additionally, it improves safety and efficiency in a dynamic use case requiring close collaboration. To achieve compliant control, our ROS2 port of a Cartesian impedance controller originally implemented by Mayr and Salt-Ducaju was utilized. A short summary of the controller implementation is provided here, for full details please refer to the original work~\cite{mayr_c_2024}.

The controller sends joint torque signals $\tau_c$, which are the sums of three components:

\begin{equation}
\tau_c=\tau_{\mathrm{c}}^{\mathrm{ca}}+\tau_{\mathrm{c}}^{\mathrm{ns}}+\tau_{\mathrm{c}}^{\mathrm{ext}}
\end{equation}

where $\tau_{\mathrm{c}}^{\mathrm{ca}}$ is the torque component commanded to achieve Cartesian impedance behavior with respect to a Cartesian reference pose, $\tau_{\mathrm{c}}^{\mathrm{ns}}$ to achieve joint impedance behavior and $\tau_{\mathrm{c}}^{\mathrm{ext}}$ to achieve the desired external force command. The Cartesian impedance component depends on the Jacobian $J(q)$, virtual Cartesian stiffness and damping matrices $K^{\mathrm{ca}}$ and $D^{\mathrm{ca}}$, and Cartesian pose error $\Delta\xi$ between reference pose and current pose:

\begin{equation}
\tau_{\mathrm{c}}^{\mathrm{ca}}=J^T(q)[-K^{\mathrm{ca}}\Delta\xi-D^{\mathrm{ca}}J(q)\dot{q}]
\end{equation}

The joint impedance component, projected in the null-space of the robot's Jacobian to not affect the Cartesian motion of the end-effector, is calculated as:

\begin{equation}
\tau_{\mathrm{c}}^{\mathrm{ns}}=(I_n-J^T(q)(J^T(q))^+)[-K^{\mathrm{ns}}(q-q^D)-D^{\mathrm{ns}}\dot{q}]
\end{equation}

where the superscript $^+$ denotes Moore-Penrose pseudoinverse matrix while $K^{\mathrm{ns}}$ and $D^{\mathrm{ns}}$ are the virtual joint stiffness and damping matrices, respectively. The torque commanded to produce the desired external force $F_{\mathrm{c}}^{\mathrm{ext}}$ is:

\begin{equation}
\tau_{\mathrm{c}}^{\mathrm{ext}}=J^T(q)F_{\mathrm{c}}^{\mathrm{ext}}
\end{equation}

The controller implementation offers multiple useful functionalities, such as updating the reference pose and Cartesian stiffness through ROS topics. These are utilized for the wrist tracking feature and to achieve compliant behavior in our setup. While the controller also allows trajectory execution, for this we instead use the basic position joint trajectory controller and present a functionality for changing controllers through voice commands and the ROS controller manager.

\subsection{Vision-based Wrist Tracking}

Vision is used to provide wrist tracking functionality, enabling the robot to track and follow the user's wrist as they handle the carbon fiber ply together. The human wrist is detected with an Intel RealSense D435 camera using the Ultralytics YOLO11n-pose model~\cite{yolo11_ultralytics}. This gives the 2D pixel of the detected wrist, which is deprojected to 3D coordinates and transformed to the robot coordinate frame using tf library~\cite{tf_foote_2013}. The detection enables the operator to be located, the distance between the human and the robot to be calculated, and for the wrist tracking based co-manipulation of the ply.

The wrist tracking co-manipulation method is implemented using the impedance controller. The reference pose of the impedance controller is updated to be the user's wrist pose with offsets in the $x$ and $y$ directions, according to the size of the ply. In our experiments, the offsets are set to \SI{0.85}{\meter} in $x$, which is slightly less than the length of the ply, and  \SI{0.375}{\meter} in $y$, equal to half the width of the ply. This allows manipulation of the ply without pulling it, which can be beneficial both for the ply and for the human operator. While similar functionality could be achieved using other control methods, such as with MoveIt servo~\cite{moveit_servo}, the Cartesian impedance controller was considered a safe and reliable option. Although both wrists are detected, only the right wrist detection is used as it is more visible and less likely to be obscured during testing.

\begin{figure*}[!b]
\centering
\subcaptionbox{\label{fig:force_without_obst}}{%
  \includegraphics[height=0.3\linewidth]{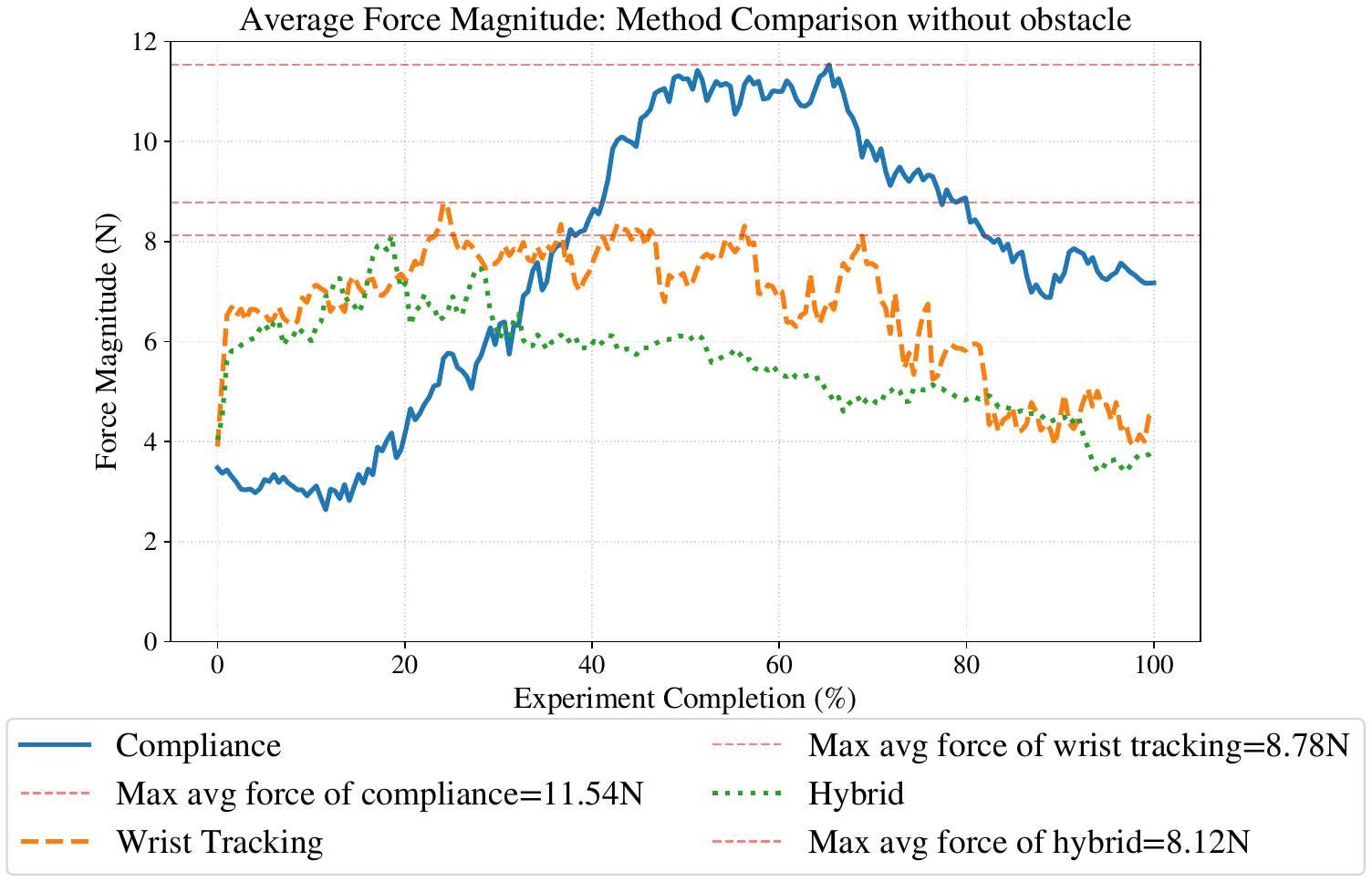} 
}\hfill
\subcaptionbox{\label{fig:force_with_obst}}{%
  \includegraphics[height=0.3\linewidth]{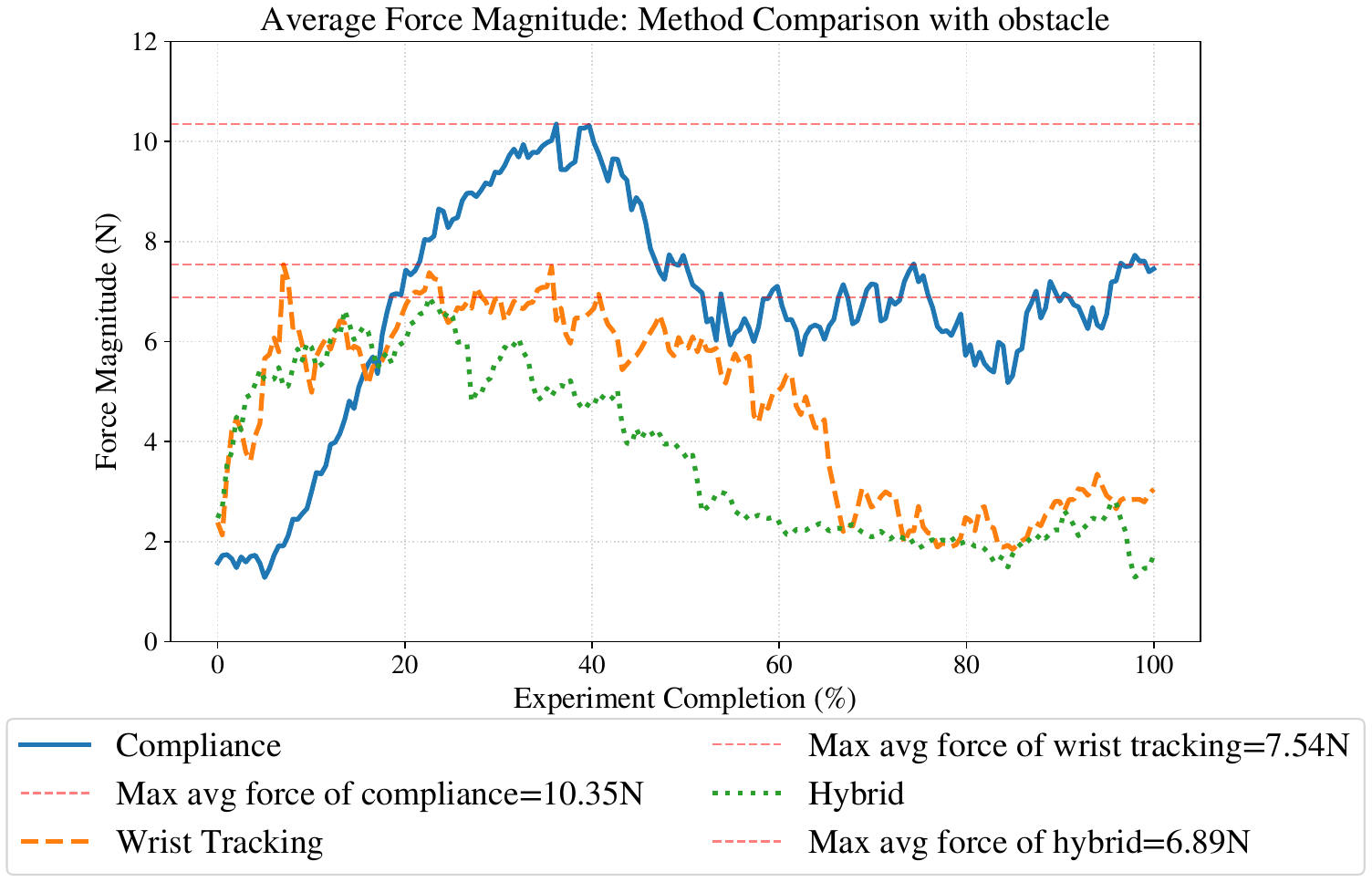} 
}\hfill
\caption{Average force norms acting on end-effector in case (a)  without obstacle and (b) with obstacle. \label{fig:force_figs}}
\end{figure*}

\begin{figure*}[!b]
\centering
\subcaptionbox{\label{fig:time_to_dist_heatmap}}{%
  \includegraphics[width=0.49\linewidth]{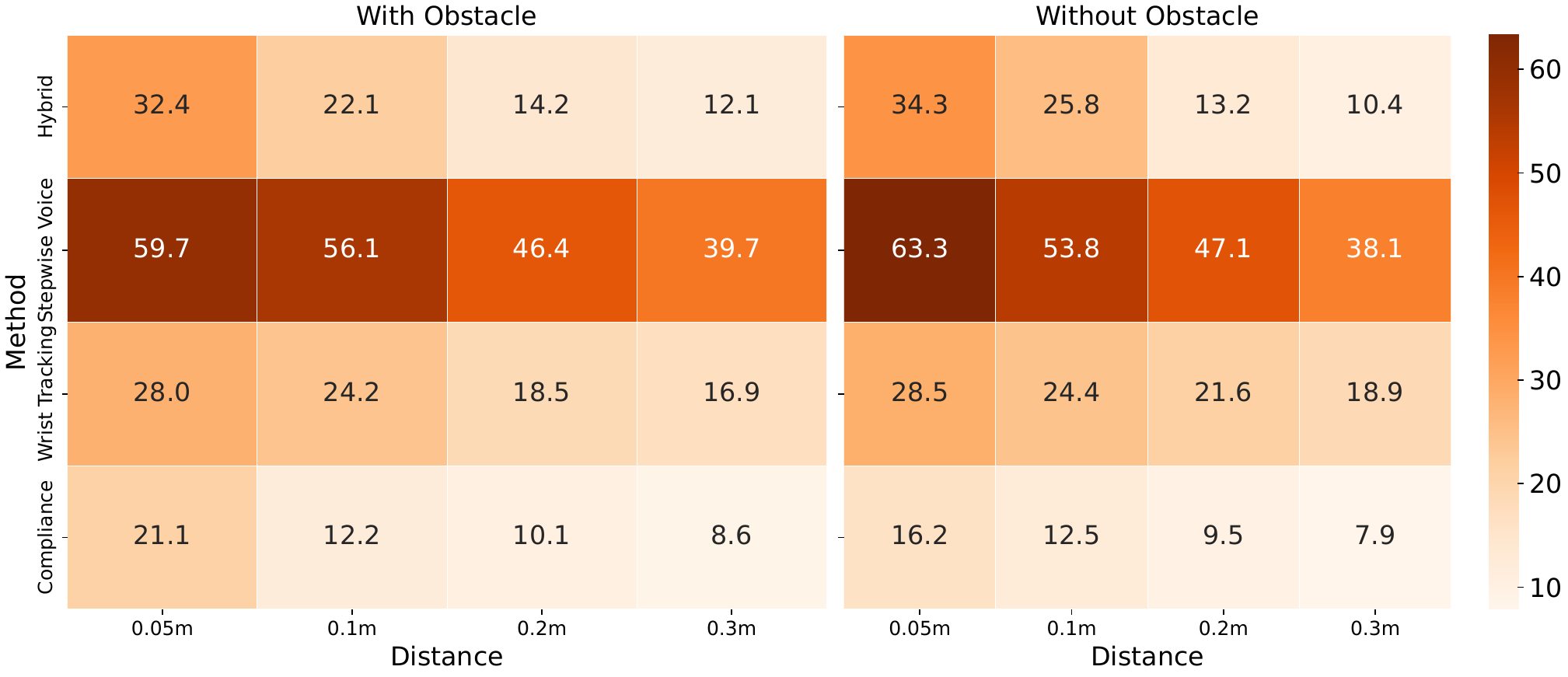} 
}\hfill
\subcaptionbox{\label{fig:paths}}{%
  \includegraphics[width=0.49\linewidth]{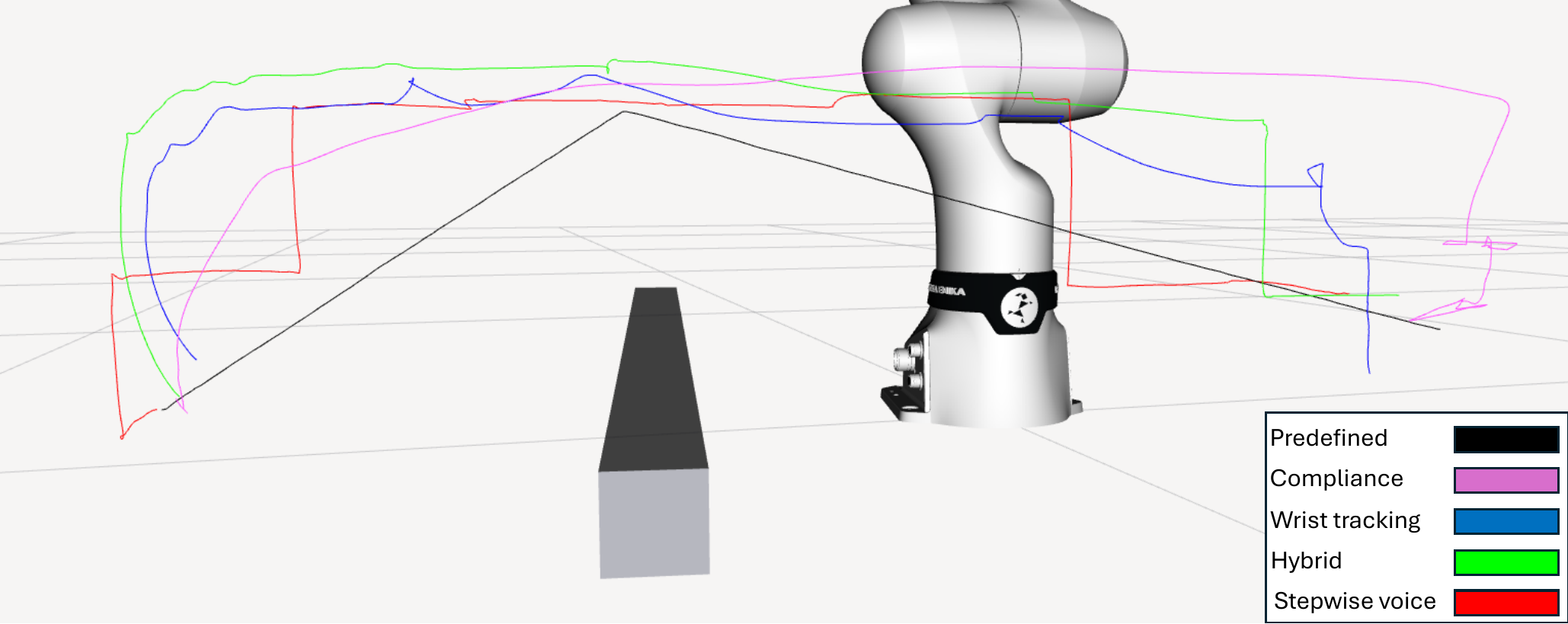} 
}\hfill
\caption{ (a) Heatmap of average times to achieve certain distance from goal. (b) Example paths to avoid obstacle from individual experiments.
\label{fig:time_and_paths}}
\end{figure*}

\subsection{Speech Perception and Communication}

Speech is an especially advantageous communication modality in co-manipulation tasks as it allows the human operator to use both hands and fully focus on the task. Additionally, the proximity of the human and the robot in close collaboration reduces the effect of environmental noise, which is a common challenge for verbal communication. Creoir EdgeVUI SDK~\cite{creoir}, which provides tools for automatic speech recognition and text-to-speech, is used in the system for human-robot and robot-human communication. It is connected to the system through ROS2 actions.

The speech interface is used to provide the operator with control of various aspects of the robot system. This includes basic movement commands, such as ``move up" or ``open gripper", enabling the whole co-manipulation process to be completed using step-by-step voice commands. In addition to basic commands for robot movement and grasping, speech is used to bring together the different modalities of the system. Through voice commands, the wrist tracking feature, controller type and compliant control can be activated or deactivated when needed. The compliant control can also be adjusted interactively by changing the directions (e.g. plane, 3D or certain direction) or levels of compliance (e.g. resisting more or less).

The TTS functionality allows the robot to give information about its status, intents and actions to the user. The success of requested actions is conveyed through messages such as ``moving home", ``closing the gripper" or ``changing controller succeeded". This kind of robot-to-human communication can enhance trust in the robot and decrease the cognitive load on the human operator.

\section{Experiments and Results}
\label{sec:experiments}

The focus of the experiments is to test different robot control methods for co-manipulation of a carbon fiber ply. Five different control methods were tested and compared both with and without an obstacle in the scene:

\begin{enumerate}
    \item Compliant control with zero stiffness
    \item Voice control using step-by-step commands
    \item Wrist tracking with robot following the user
    \item Robot has a predefined trajectory while user follows
    \item A hybrid approach combining wrist tracking, stepwise voice commands and compliance
\end{enumerate}

The experimental procedure consists of the user and robot picking up a sheet of carbon fiber, maneuvering it over a mold while avoiding the possible obstacle, and placing the ply on the mold. The procedure is repeated three times for each control strategy, with consistent behavior seen across trials. Given the focus is on the transportation phase of the process, the data presented is from the moment the robot starts moving until it reaches and stabilizes at the final placement point, excluding the gripper opening and the robot moving home. This minimizes effects independent of the control method, such as delays in speech commands.

\begin{figure*}[htbp]
\centering
\subcaptionbox{\label{fig:compliance_1}}{%
  \includegraphics[height=0.18\linewidth]{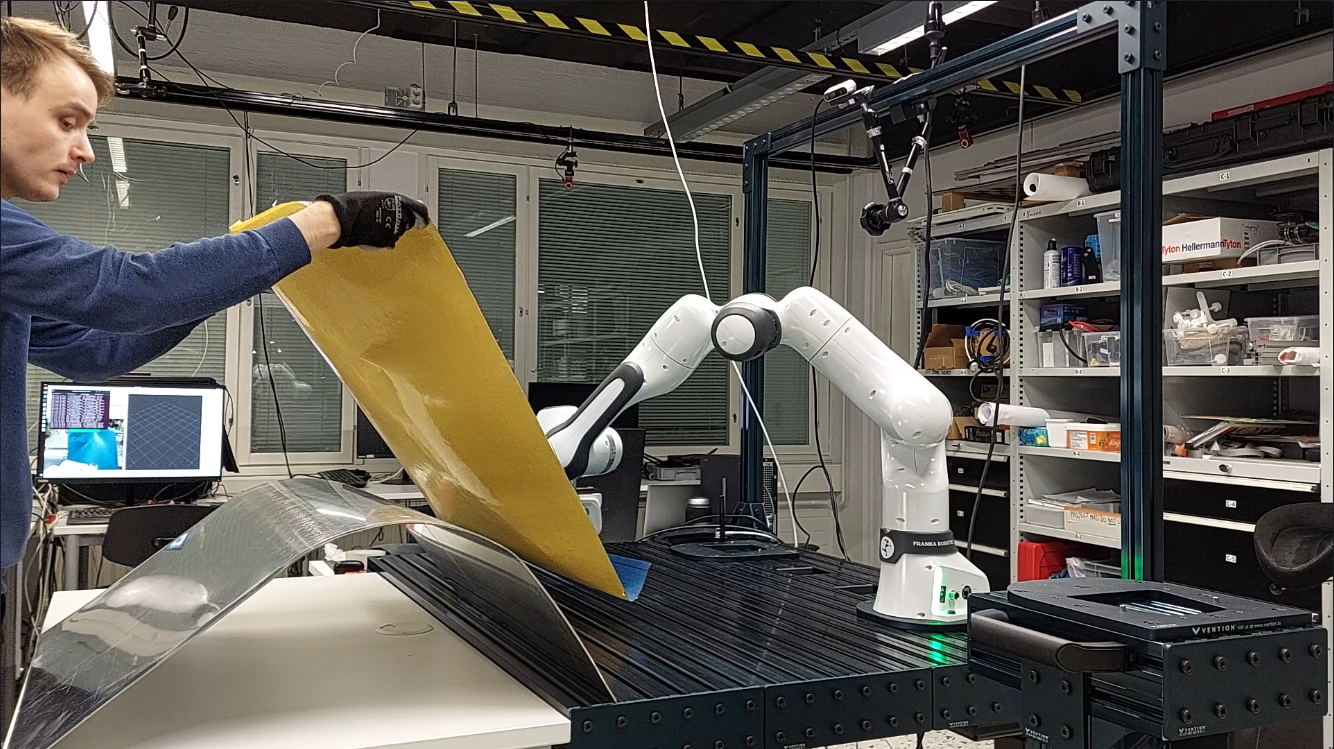} 
}\hfill
\subcaptionbox{\label{fig:compliance_2}}{%
  \includegraphics[height=0.18\linewidth]{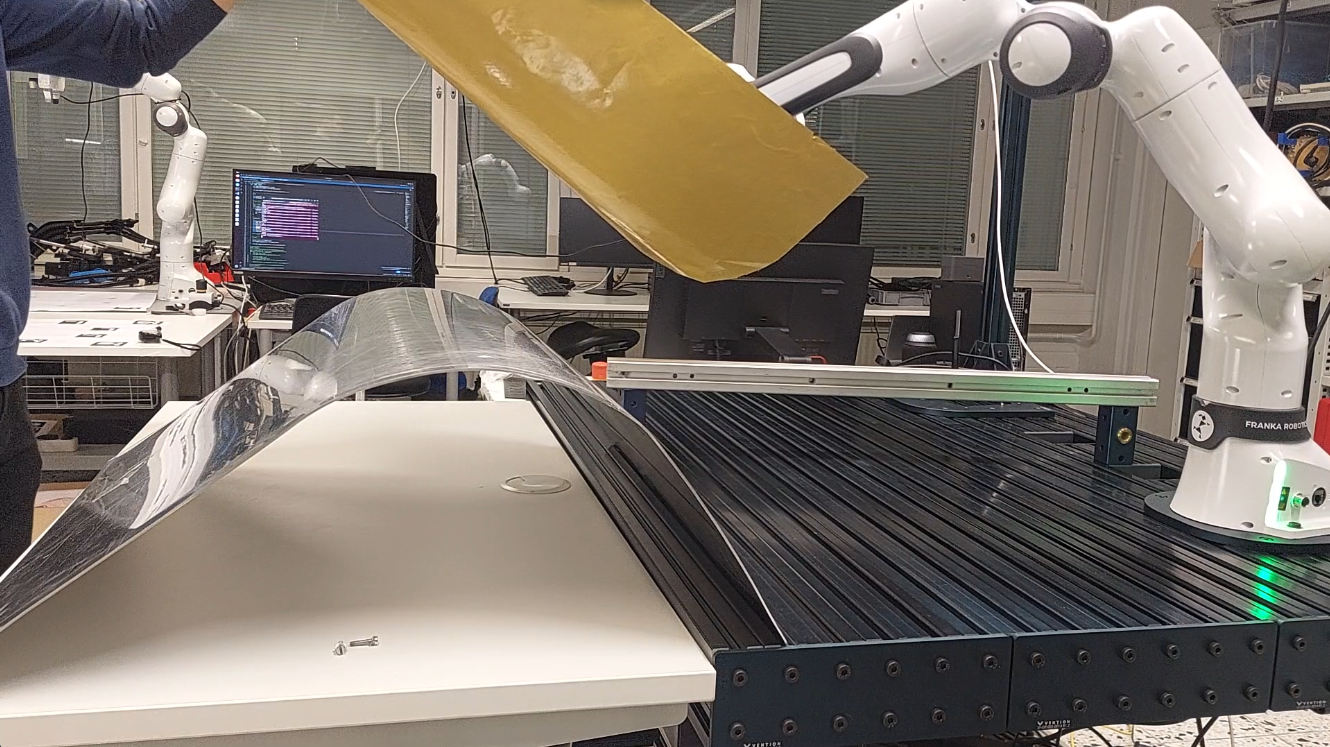} 
}
\hfill
\subcaptionbox{\label{fig:compliance_3}}{%
  \includegraphics[height=0.18\linewidth]{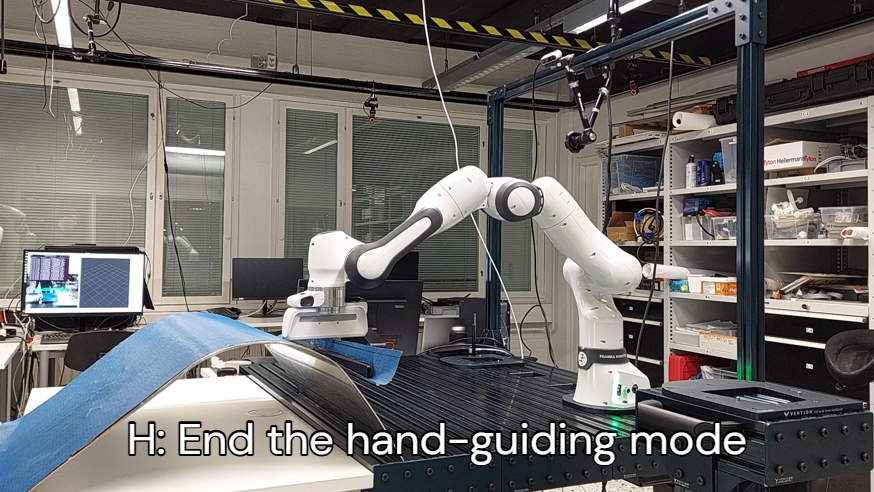} 
}\hfill
\caption{Examples from compliant control trials. (a) Without the obstacle, mostly planar movement is sufficient to approach the mold. (b) Avoiding the obstacle requires 3D movement. (c) Compliant control can be ended through speech.\label{fig:compliance_examples}}
\end{figure*}

\begin{figure*}[htbp]
\centering
\subcaptionbox{\label{fig:stepwise_1}}{%
  \includegraphics[height=0.18\linewidth]{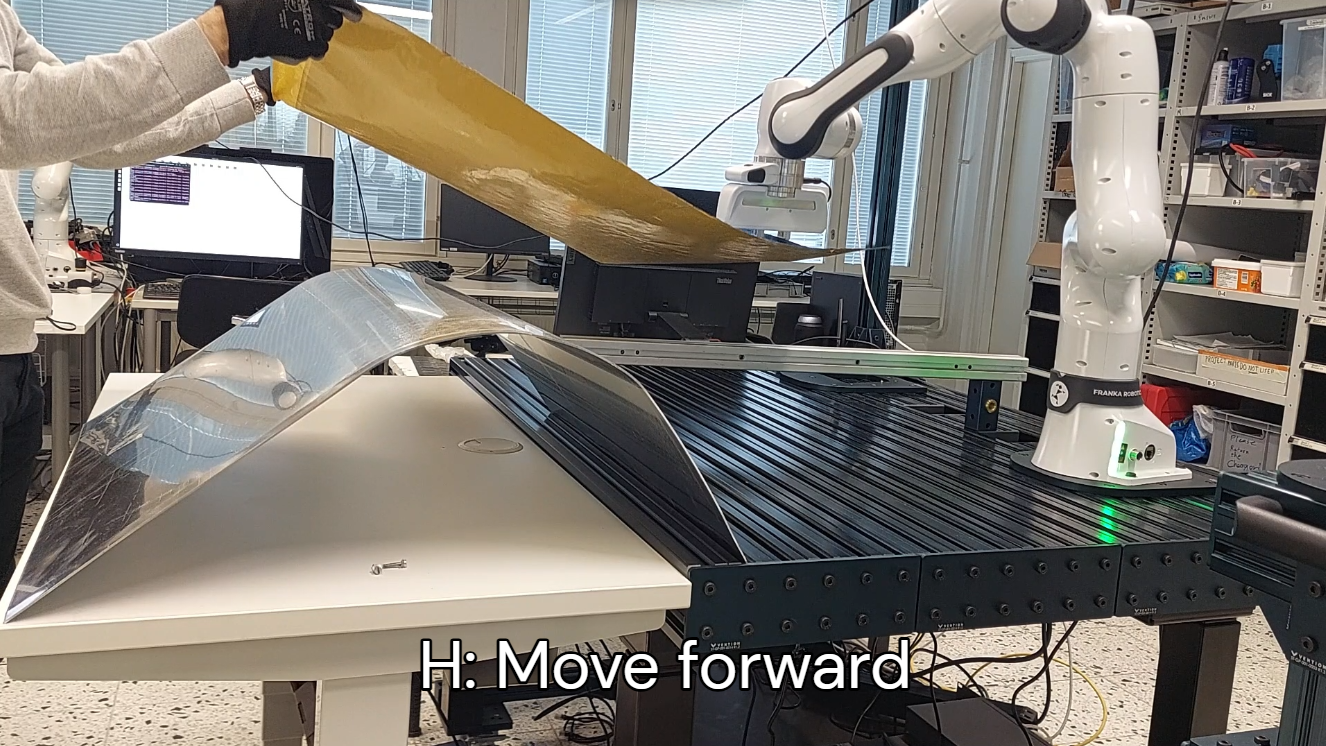} 
}\hfill
\subcaptionbox{\label{fig:stepwise_2}}{%
  \includegraphics[height=0.18\linewidth]{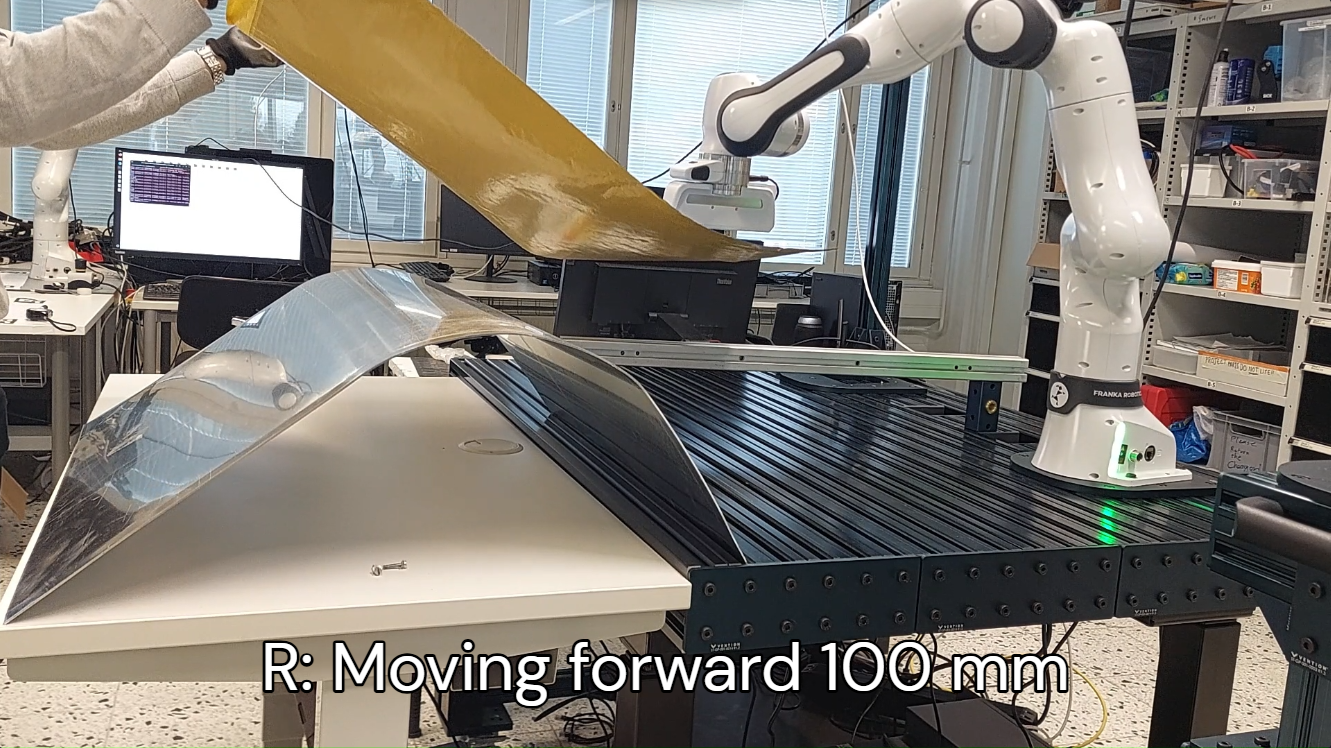} 
}
\hfill
\subcaptionbox{\label{fig:stepwise_3}}{%
  \includegraphics[height=0.18\linewidth]{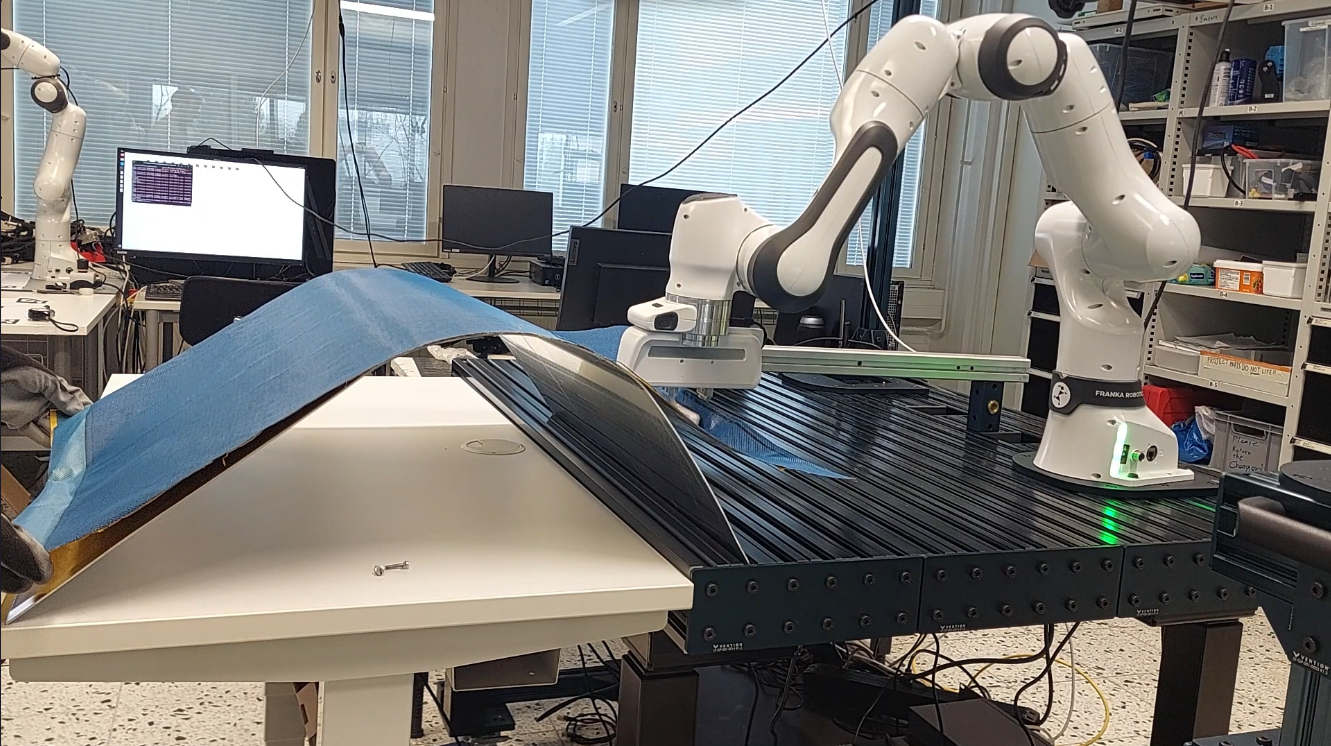} 
}\hfill
\caption{Examples of (a) stepwise voice commands with (b) robot feedback, enabling the robot to be moved in all directions. \label{fig:stepwise_voice_examples}}
\end{figure*}

\begin{figure*}[htbp]
\subcaptionbox{\label{fig:wrist_tracking_1}}{%
  \includegraphics[height=0.18\linewidth]{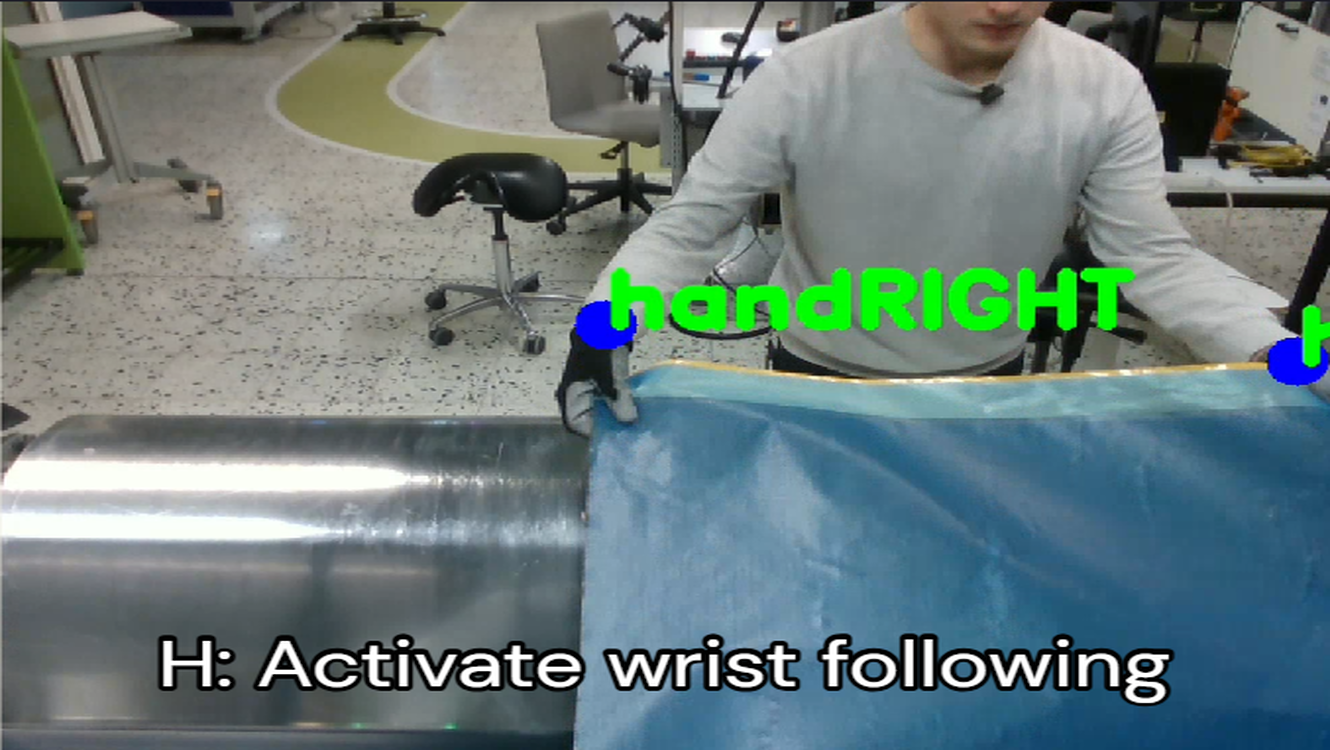} 
}\hfill
\subcaptionbox{\label{fig:wrist_tracking_2}}{%
  \includegraphics[height=0.18\linewidth]{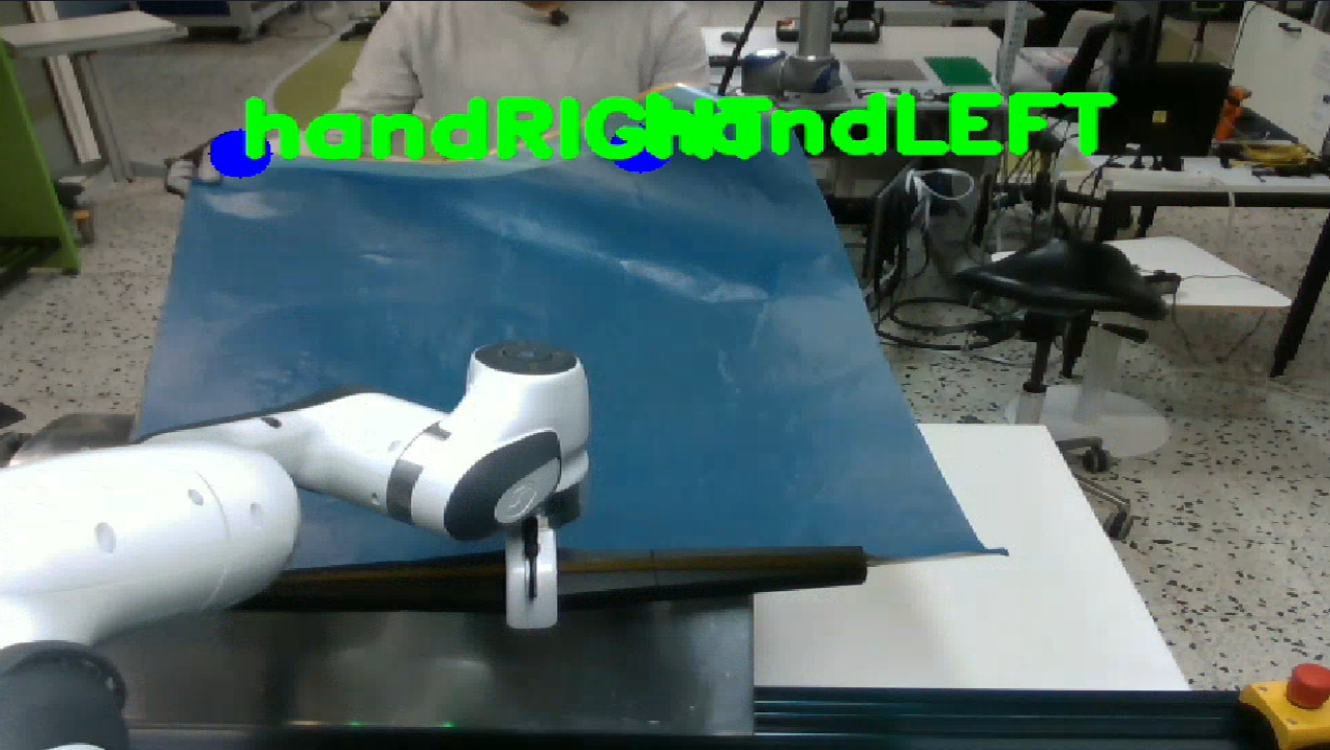} 
}
\hfill
\subcaptionbox{\label{fig:wrist_tracking_3}}{%
  \includegraphics[height=0.18\linewidth]{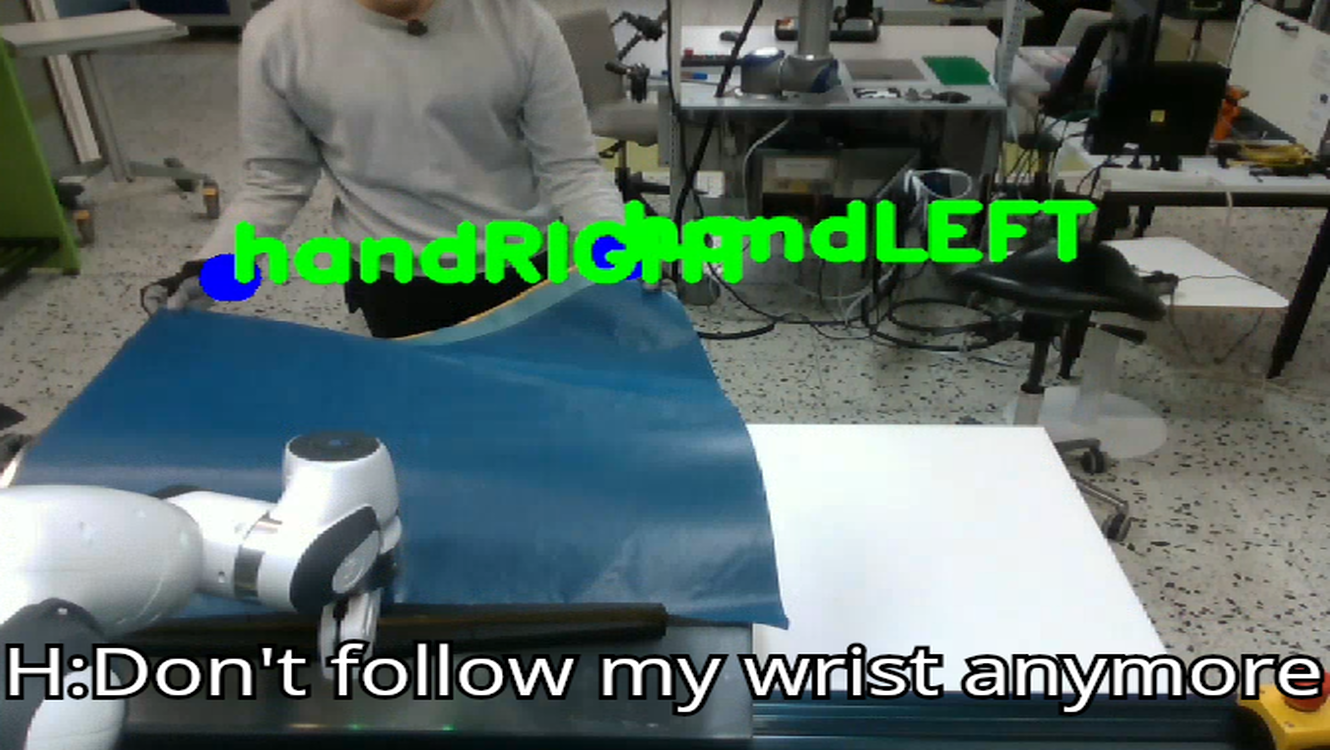} 
}
\caption{Snapshots of wrist tracking: (a) the user activates wrist tracking through a voice command, (b) the robot follows the human, and (c) the user ends it through a voice command.\label{fig:wrist_tracking_examples}}
\end{figure*}

\begin{figure*}[htbp]
\subcaptionbox{\label{fig:combination_1}}{%
  \includegraphics[height=0.18\linewidth]{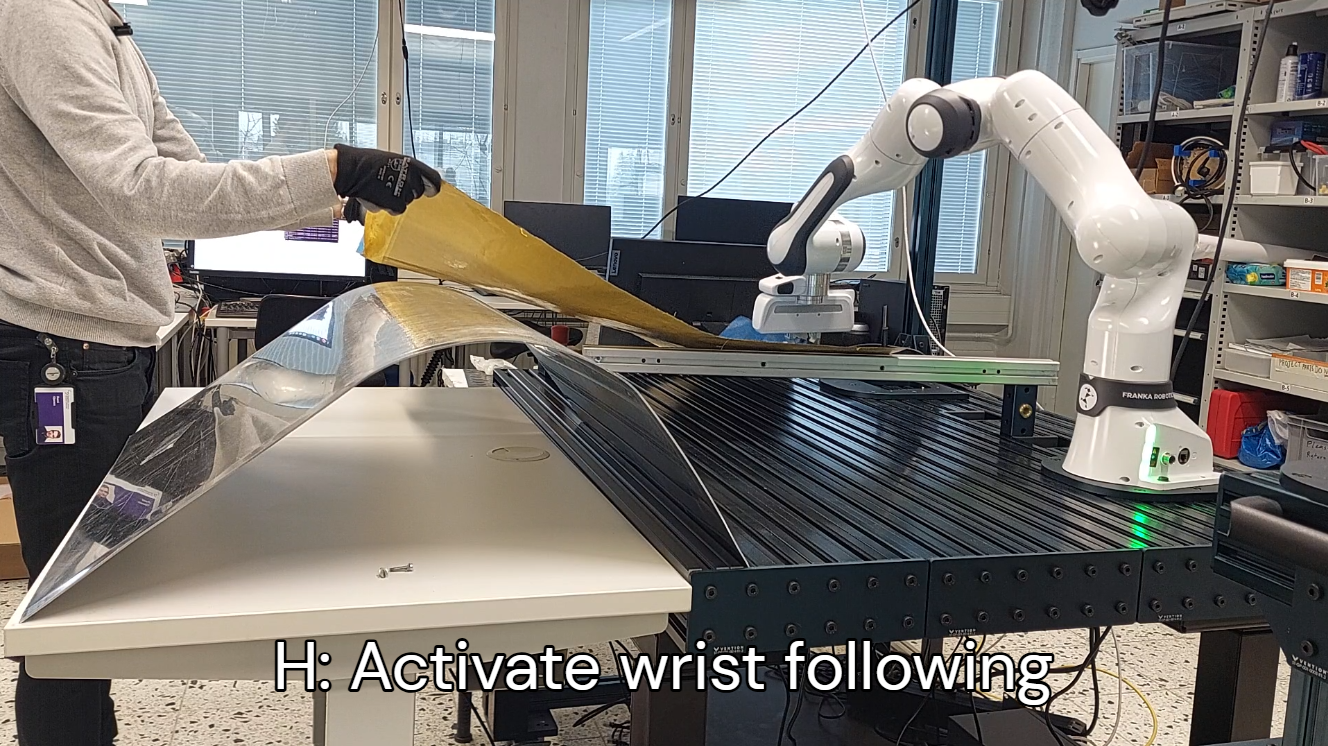} 
}\hfill
\subcaptionbox{\label{fig:combination_2}}{%
  \includegraphics[height=0.18\linewidth]{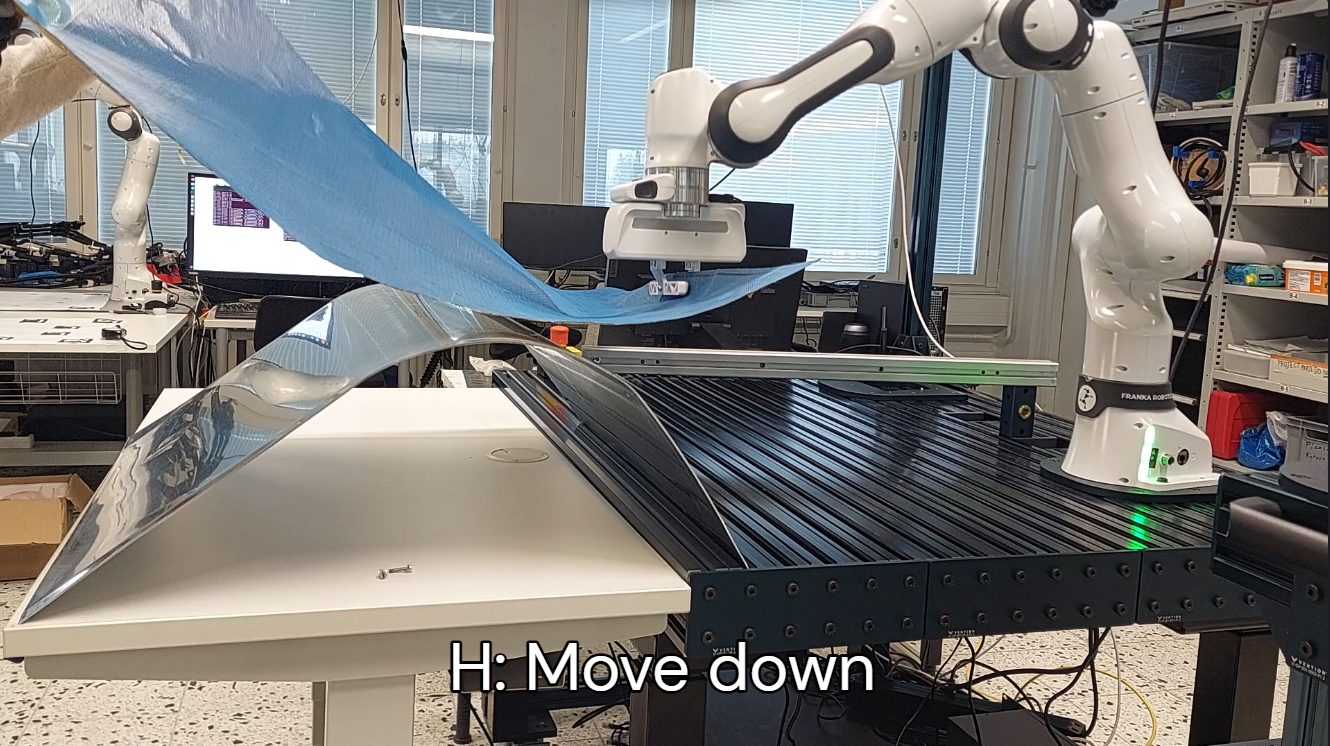} 
}
\hfill
\subcaptionbox{\label{fig:combination_3}}{%
  \includegraphics[height=0.18\linewidth]{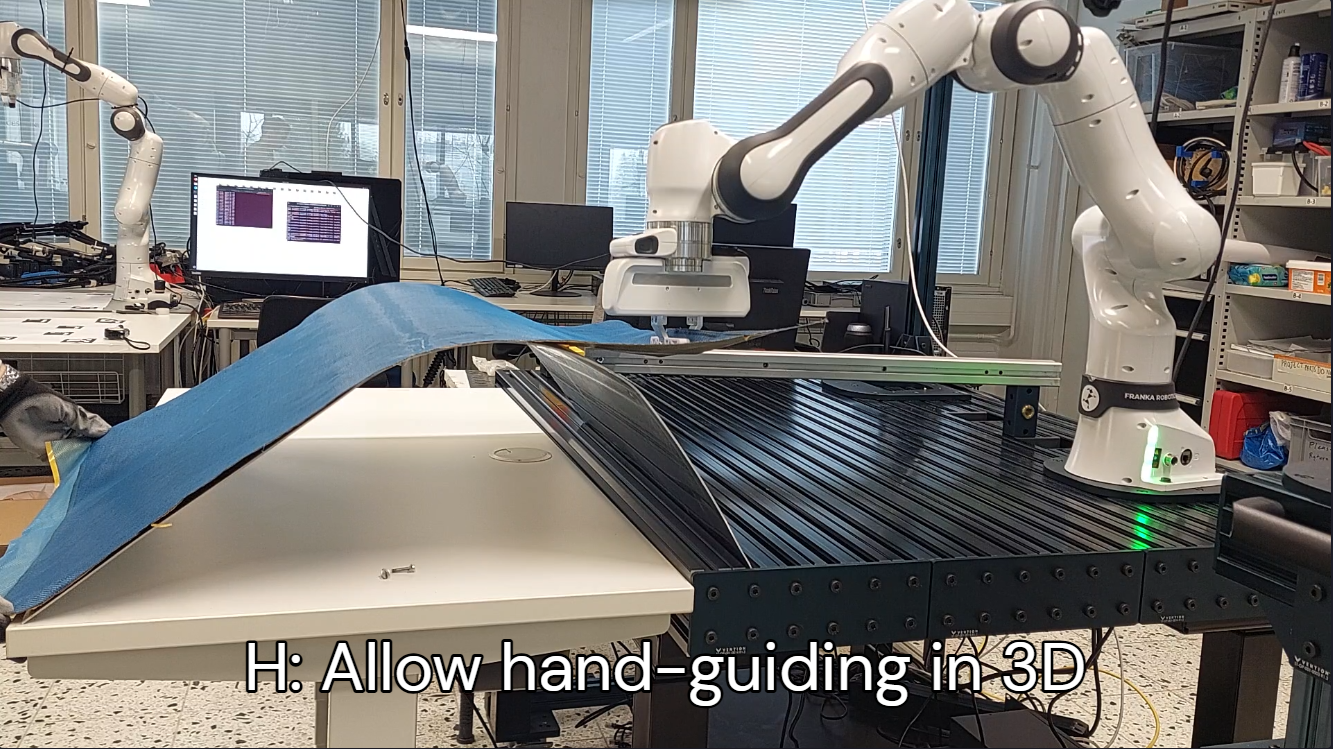} 
}
\caption{Phases of the hybrid approach: (a) the user first moves the robot close to the placement point with wrist tracking, (b) adjustments are made through voice commands, (c) final hand guiding with compliant control. \label{fig:combination_examples}}
\end{figure*}

To compare the different control strategies, various metrics were recorded during the trial.
Fig.~\ref{fig:force_figs} presents the average force on the robot end-effector across the different trial types. Fig.~\ref{fig:time_to_dist_heatmap} shows the time to reach different distance tolerances around the final end position. Table~\ref{tab:pros_and_cons_table} presents overall average trial times and path lengths, along with a comparison of pros and cons for the different methods.
The force data are presented for the compliance control method, since it relies on pulling the material, and both the wrist tracking and hybrid methods for comparison. The non-zero force norm during wrist tracking in Fig.~\ref{fig:force_figs} is caused by the movement of the robot, not the user pulling. 
Additionally, example paths to avoid the obstacle with different methods are presented in Fig.~\ref{fig:paths}. Examples of the experiments performed can be found in the attached repository and in Figs.~\ref{fig:compliance_examples}--\ref{fig:combination_examples}. In the following subsections, the results of the different control methods are discussed in further detail.

\subsection{Compliant Control}

Voice commands can be used to adjust the behavior of the robot by changing whether it is compliant or not, how compliant it is (e.g. stiffness) and the directions of compliance (e.g. forward, planar or 3D). In the experiments, compliance with
zero stiffness was tested. Fully compliant robot behavior allows the human operator to safely and easily control the manipulation process by pulling or moving the ply while the robot follows. As can be seen from Table~\ref{tab:pros_and_cons_table}, it is also the fastest method, other than using a predefined trajectory. Including stiffness could improve placement accuracy. However, it would require a greater force when pulling, which could damage the material and put a strain on the human operator, making it unsuitable for a long transportation task. Given the requirement for tension in the material to pull the robot, a major drawback for compliance is the difficulty in moving the ply backwards.

By changing the directions of compliance, the robot can better adjust to different scenarios and surfaces. For example, since the surface where the ply is placed is curved, or when there are obstacles to be avoided during the manipulation process, as in the second case of our experiments, full compliance in 3D is beneficial. Meanwhile, full compliance on a plane can improve the placement accuracy and controllability on planar surfaces by reducing end-effector floating behavior.

\subsection{Following a Predefined Trajectory}
Preprogramming the robot to follow a predefined trajectory while the human adapts is the fastest way to perform the manipulation process and also does not require any additional physical effort from the user. In dynamic manufacturing environments with varying processes, a fully predefined approach may not be possible and has significant drawbacks in terms of flexibility. The required reprogramming of the robot presents significant extra work and effort every time the environment, ply size or placement point changes.

\subsection{Stepwise Voice Control}
The voice control method enabled the user to control the robot to complete the task. On average, 12 discrete commands were used by the user to complete the task. In this case, a predefined step size of 10 centimeters was used for the movement, even though the distance could also be specified as a part of the command, for example, ``Move 20 centimeters forward". This approach was chosen to limit the effect of learning and over-optimization of the distances. An example sequence of commands would be:

\begin{itemize}
    \item ``Move forward"
    \item ``Move right"
    \item ``Move up"
    \item ``Move forward"
\end{itemize}

Moving the robot through voice commands presented an intuitive way to control the robot for adaptive co-manipulation tasks. It can also better adapt to changes in the process or environment compared to the predefined trajectory method. This method, however, led to bumpy, stop-start movement, resulting in a long path and slow task execution, as shown in Fig.~\ref{fig:time_and_paths} and Table~\ref{tab:pros_and_cons_table}.
The need to either use predefined steps or predict the distance required and specify it as part of the command means the method has limited accuracy when finely positioning the ply.

\subsection{Wrist Tracking}

Wrist tracking presents a multimodal method for controlling the robot in the co-manipulation task. Based on the detected wrist pose and including offsets based on the ply size, the reference pose of the impedance controller is constantly updated, resulting in the robot following the human movements. As the human moves to place the ply over the mold, the robot therefore follows at a constant distance apart.

One advantage of this method compared to compliance is that it is easier to move the ply upwards, backwards or sideways without causing the ply to get twisted. This method also requires no pulling, which can be advantageous for both the ply and the human, while allowing smoother collaboration compared to the robot moving in steps or in a fully predefined manner. However, further development is required to ensure easier and more accurate control. Learning to use this control method effectively takes time, while accurate placement of the ply, especially close to the surface can also be challenging. The need for an external camera also presents the risk of occlusions.

\subsection{Hybrid Approach}
The hybrid approach offers an example of combining different control methods during runtime. The user first moves the robot close to the placing point with wrist tracking, before making adjustments through voice commands and hand guiding. This approach resulted in the shortest end-effector path, was faster than using voice commands alone and caused a smaller force on the end-effector compared to only using compliance. The additional commands and time when changing the control method increase the duration of the task for this approach, as can be seen from the jump in time between distances of 0.2\,m and 0.1\,m in Fig.~\ref{fig:time_to_dist_heatmap}. The total time is still comparable to other methods. 

\begin{table*}[t]
\centering
\caption{Pros, cons, average total times and path lengths of different control methods}
\label{tab:pros_and_cons_table}
\small
\begin{tabularx}{\textwidth}{lXXccc}
\toprule
\textbf{Control method} & \textbf{Pros} & \textbf{Cons} & \textbf{Average time (s)} & \textbf{Path length (m)} \\
\midrule
\textit{Compliance} &
Intuitive and safe control of the process; human-led interaction; adaptable to different scenarios by changing the directions and level of compliance &
Requires force which can be harmful for the ply and operator; difficult to move the ply accurately in 3D &
\makecell[ct]{Without obstacle: 25.3;\\ With obstacle: 31.1} & \makecell[ct]{Without obstacle: 1.02;\\ With obstacle: 1.16} \\
\addlinespace

\textit{Predefined trajectory} &
Fast and straightforward; requires no force and less human effort; consistent process time &
Not human-centric; human not in full control (trust, safety); not suitable in dynamic environments; requires rework if environment changes &
\makecell[ct]{Without obstacle: 10.4;\\ With obstacle: 11.5} & \makecell[ct]{Without obstacle: 0.84;\\ With obstacle: 0.90} \\
\addlinespace

\textit{Stepwise voice control} &
Intuitive; easy movement in all directions; adaptable to new scenarios; human in control &
Bumpy movement and discontinuous collaboration; difficult to estimate accurate distances; inaccurate commands reduce fluency &
\makecell[ct]{Without obstacle: 64.7;\\ With obstacle: 60.7} & \makecell[ct]{Without obstacle: 1.26;\\ With obstacle: 1.37} \\
\addlinespace

\textit{Wrist tracking} &
No force required; easy to move in 3D &
Limited placement/control accuracy; more difficult to control than compliance; possible safety issues in unexpected situations &
\makecell[ct]{Without obstacle: 33.6;\\ With obstacle: 37.1} & \makecell[ct]{Without obstacle: 1.13;\\ With obstacle: 1.16} \\
\addlinespace

\textit{Hybrid} &
Combines advantages of different methods; requires less force than compliance alone; faster than stepwise voice commands alone &
Requires high number of commands and switching control methods which may extend task completion time and complexity  &
\makecell[ct]{Without obstacle: 36.2;\\ With obstacle: 39.5} & \makecell[ct]{Without obstacle: 0.99;\\ With obstacle: 1.04} \\
\bottomrule
\end{tabularx}
\end{table*}

\section{Discussion}
\label{sec:discussion}

The results from the experiments found that, while the predefined trajectory can be fast and stepwise voice commands intuitive, compliance offered the human operator more control over the process with stable and adaptable co-manipulation. Adjusting the compliance level and directions accordingly can increase the adaptability of the robot in different scenarios and improve the accuracy of the process. Incorporating vision and wrist tracking allowed the robot to be controlled without pulling the ply, which is advantageous for both the human operator and the material being manipulated, while also allowing easier control in 3D and directions that are difficult with compliance. A hybrid approach allows combining the advantages of wrist tracking, stepwise voice commands and compliance. A summary of the advantages, disadvantages, average total times and path lengths of the different methods is presented in Table \ref{tab:pros_and_cons_table}.

Force and vision are commonly used modalities in co-manipulation literature. In most cases however, vision is either not used for controlling the robot at all, or it is used for correcting the control or trajectory of the robot. In our wrist-tracking approach, vision is directly involved by producing a reference pose for the robot. Compared to the work of De Schepper et al.~\cite{de_schepper_towards_2021}, which only uses 2D RGB image data and can therefore only move the robot vertically or horizontally, our implementation allows movement towards or away from the human based on skeleton tracking. We regard the ability to move away from the human as an important feature, especially given that direction is difficult to achieve through methods based on compliance or force.
The model of Bonci et al.~\cite{bonci_human-robot_2024} includes certain simplifications and assumptions, and lacks real-world experiments, having only been tested in simulation.

While the advantages of speech are well-known in HRC, the number of examples in co-manipulation cases is limited. We present an approach to integrate voice, vision and force to create a multimodal system for co-manipulation of carbon fiber plies. Additionally, compared to model- or learning- based methods~\cite{makris_deformable_2022,8769898}, our approach does not require modeling the ply, which can be a complex and computationally heavy operation.
 
There are certain limitations in our work. In addition to those related to the use case and setup mentioned in Section~\ref{sec:use_case_description}, there are limitations in the control methods. Compliance can be challenging to use and accurately control in 3D as it requires pulling on the ply diagonally upward, instead of maintaining the preferred planar form of the material. As mentioned previously, moving the ply backwards can also be challenging without twisting the ply, while any end-effector floating behavior from imperfect gravity compensation reduces the placement accuracy. With stepwise voice commands, the accuracy and fluency of collaboration depend on the operator's ability to estimate distances and give appropriate commands. Meanwhile, the wrist tracking feature requires further improvements in accuracy, fluency and safety, for example in cases with wrong detections, unexpected operator movement or multiple people in the workspace.

This paper presents isolated investigations used to analyze the pros and cons of different control methods for the transportation phase of collaborative carbon fiber sheet layup. Future work will be to integrate the methods into a complete industrial scenario. Given the various pros and cons discussed, it is clear that a combination of the control methods presented offers a likely solution to the problem. For example, the robot could be moved close to the placement location using either the wrist tracking feature or a predefined trajectory, before accurate adjustments with voice commands and compliance. The wrist detection functionality could also be used for safety-related features, such as adjusting robot speed or control mode based on human location and distance between human and robot.

\section{Conclusion}
\label{sec:conclusion}

This work proposed how a multimodal combination of speech, vision and force could be used in human-robot co-manipulation of a carbon fiber ply. Different control methods, including compliance, a predefined trajectory, stepwise voice commands, wrist tracking and a hybrid method, were tested and compared. The results suggest that while the predefined trajectory method can be fast and stepwise voice commands intuitive, compliance offers the operator greater and more stable control of the process. Changing the levels and directions of the compliance accordingly can improve the adaptability of the robot in different scenarios and improve the accuracy of the process. Meanwhile, incorporating vision and wrist tracking allows the robot to be controlled without pulling on the ply, which is advantageous both for the operator's ergonomics and the ply. A hybrid approach allows the advantages of different methods to be combined.
While this paper presents preliminary developments and a comparison of methods focused on the transportation phase, future work will focus on implementing the complete industrial use case.

\section*{Acknowledgments}
This project has received funding from the European Union's Horizon Europe research and innovation program under grant agreement no. 101135708 (JARVIS). 
The authors would like to thank Ossi Parikka for his contributions to the development of the control architecture and the ROS2 implementation of the cartesian impedance controller.

\bibliographystyle{IEEEtran}
\bibliography{IEEEabrv,refs}


\end{document}